\documentclass[11pt,a4paper]{article}
\usepackage{times,latexsym}
\usepackage{url}
\usepackage[T1]{fontenc}

\usepackage[acceptedWithA]{tacl2021v1}
\usepackage{xspace,mfirstuc,tabulary}

\newif\iftaclinstructions
\taclinstructionsfalse % AUTHORS: do NOT set this to true
\iftaclinstructions
\renewcommand{\confidential}{}
\renewcommand{\anonsubtext}{(No author info supplied here, for consistency with
TACL-submission anonymization requirements)}
\newcommand{\instr}
\fi

\iftaclpubformat % this "if" is set by the choice of options

\else

\fi

\usepackage{amsmath}
\usepackage{paralist}
\setdefaultleftmargin{0em}{}{}{}{}{}

\newcommand{\textnl}[1]{`\textsl{#1}'}
\newcommand{\sref}[1]{Section~\ref{#1}}
\newcommand{\exref}[1]{Example~\ref{#1}}

\usepackage{pgfplots}
\pgfplotsset{width=8cm,compat=1.9}
\usepackage{amsfonts}
\usepackage{expex}

\usepackage{multirow}
\usepackage{rotating}

\title{Shared Lexical Items as Triggers of Code Switching}
\author{Shuly Wintner$^\dagger$ 
    \and 
    Safaa Shehadi$^\dagger$ 
    \and Yuli Zeira$^\dagger$ 
    \and Doreen Osmelak$^\diamond$ 
    \and Yuval Nov$^\ddagger$ \\
    $^\dagger$Department of Computer Science, 
    University of Haifa, Israel\\
    $^\diamond$Department of Language Science and Technology, 
    Universit\"at des Saarlandes, Germany\\
    $^\ddagger$School of Public Health, 
    University of Haifa, Israel\\
\texttt{shuly@cs.haifa.ac.il},
\texttt{safa.shehadi@gmail.com}, \texttt{yuli.zeira@gmail.com}\\
\texttt{s9doosme@stud.uni-saarland.de},
\texttt{yuval@stat.haifa.ac.il}
    }

\begin{document}
\maketitle

\begin{abstract}
Why do bilingual speakers code-switch (mix their two languages)? Among the several theories that attempt to explain this natural and ubiquitous phenomenon, the \emph{Triggering Hypothesis} relates code-switching to the presence of lexical triggers, specifically cognates and proper names, adjacent to the switch point. We provide a fuller, more nuanced and refined exploration of the triggering hypothesis, based on five large datasets in three language pairs, reflecting both spoken and written bilingual interactions. Our results show that words that are assumed to reside in a mental lexicon shared by both languages indeed trigger code-switching; that the tendency to switch depends on the distance of the trigger from the switch point; and on whether the trigger precedes or succeeds the switch; but not on the etymology of the trigger words. We thus provide strong, robust, evidence-based confirmation to several hypotheses on the relationships between lexical triggers and code-switching.
\end{abstract}

\section{Introduction}
More than half the world’s population today is multilingual, yet our understanding of the underlying linguistic and cognitive principles that govern multilingual language is imperfect. 
It is largely based on controlled laboratory studies, and only recently have psycholinguists begun exploring the extent to which insights from laboratory experiments can be applied in a real-world, communicative setting
\citep{VALDESKROFF2023}.
Lacking firm theoretical underpinnings, contemporary
language technology often does not reflect the ubiquity of multilingual
communication. 

We focus in this paper on \emph{code-switching (CS)}, the natural tendency of bilingual speakers conversing with each other to switch between two languages, sometimes within a single utterance or even a single word. 
Our main goal is to explore a specific hypothesis related to CS, namely that certain words tend to \emph{trigger} CS more than others.
The main contribution of this work is theoretical, but we trust that its results will be instrumental for improving future multilingual NLP applications. 

Several competing theories try to explain CS, and in particular to identify the factors that contribute to the (typically unconscious) decision of a speaker to code-switch. 
Speakers are conjectured to code-switch when the concept they are about to utter is more \emph{accessible} in the other language \citep{heredia2001bilingual}; 
or more \emph{specific}, lacking precise enough words in the current language \citep{Backus:2001}; 
or carrying a major \emph{information} load, so that the switch signals to the listener that an important concept is introduced \citep{myslin2015code}. 
The tendency to code-switch is influenced by linguistic factors (e.g., \emph{cognates} are assumed to trigger CS), socio-linguistic factors (e.g., the fluency of the interlocutors in each of the two languages), demographic ones (e.g., the age, gender or provenance of dialogue participants), and more \citep{Myers-Scotton:1993,Myers-Scotton:1998,auer:1998,Nilep:2006}. 

We focus on the \emph{triggering hypothesis}, whereby ``lexical items that can be identified as being part of more than one language for the speaker [...] may facilitate a transversion from one language to another'' \citep[p.~162]{Clyne:2003}.
This hypothesis was explored extensively in the past, but earlier studies were limited in scope, were based on limited data, and addressed only spoken language.

This work makes several contributions. First,
we investigate a specific type of lexical trigger: we define a category of lexical items (mainly proper names and culturally specific terms) that we expect to reside in more than one (or alternatively, in a \emph{shared}) mental lexicon (\sref{sec:goals}). We also pay attention to whether such items originate in one of the two languages or in a third language.
Second, unlike previous work, which dealt exclusively with spoken data, we investigate both spoken and written data, in five large datasets that include CS in three language pairs: English--Spanish, English--German and English--Arabic%
\footnote{The Arabic in this work reflects mostly the dialects of Egypt and Lebanon, and is written in \emph{Arabizi}, an informal writing system that uses the Roman alphabet. See Section~\ref{sec:data}.}
(\sref{sec:data}).
Third, while we employ the same statistical test that has been used by previous works to assess the association between such shared items and CS, we augment the analysis by also quantifying the \emph{magnitude} of this association as an indication of the strength of the phenomena we observe (\sref{sec:methodology}), thereby adding statistical rigor to our analysis.

Our results (\sref{sec:results}) show strong associations between the presence of shared items (the type of trigger we focus on) and the tendency to code-switch, in all language pairs and datasets. We also provide a thorough and nuanced analysis (\sref{sec:analysis}) of the location of the shared item with respect to the switch point, showing that the tendency to switch is lower when the trigger is adjacent to the switch rather than precedes it; and that the association between triggers and CS diminishes as the shared items are more distant from the switch point.
Overall, we provide a much fuller, more nuanced picture of the relationships between lexical triggers and CS than was available so far.%
\footnote{All resources produced in this work, including the annotated datasets and the code, are publicly available on our \href{https://github.com/HaifaCLG/Triggering}{GitHub repository}.}

\section{Related work}
\label{sec:related-work}
Multilinguality is becoming more and more ubiquitous, to the extent that
psycholinguists increasingly
acknowledge that bilingualism is the rule and not the exception
\citep{Harris:Nelson:1992}.
\citet[page~16]{Grosjean:2010} stated that ``bilingualism is a worldwide phenomenon, found on all continents and in the majority of the countries of the world''
and \citet{grosjean2012psycholinguistics} assessed that
more than half the world’s population today is multilingual.

Monolingual and multilingual speakers
alike seamlessly adjust their communication style to their
interlocutors
\citep{bell1984language,pickering2004toward,kootstra2012priming,gallois2015communication,fricke2016primed}.
Specifically, when interlocutors share more than one language, they almost
inevitably engage in CS
\citep{sankoff1981formal,muysken2000bilingual,Clyne:2003}.

Most linguistic research on CS has focused
on \emph{spoken} language \citep[\textit{inter
  alia}]{lyu2010,LI2014,deuchar2014building}.  However, with the rise
of social media, \emph{written} CS
\citep{sebba2012language} has become a pervasive
communication style \citep{Rijhwani2017}.  The spoken language domain
is not directly comparable to the written one, and findings on CS in
written conversations differ somewhat from those in speech
\citep{mcclure2001oral,chan2009english,gardner2015code}.
The work we present here addresses both modalities.

Various competing theories attempt to explain CS, or at least to propose factors that contribute to the tendency of bilingual speakers to code-switch. 
Notable among them is the \emph{triggering hypothesis}, 
which states that specific lexical items that may be included in more than one mental lexicon for the speaker \emph{trigger} switching \citep{Clyne:2003}.
Such lexical items include, according to \citeauthor{Clyne:2003}, \emph{lexical transfers} (i.e., borrowed words and expressions), \emph{bilingual homophones} (including loans from a third language) and \emph{proper nouns}. 
In this work we focus on a specific type of potential triggers, consisting mainly of proper names but including also culturally specific lexical items that originate in one language and do not have a readily available translation in the other language (e.g., \textnl{taco}, originally from Spanish, in English--Spanish dialogues, or \textnl{muezzin}, originally from Arabic, in English--Arabic conversations).

The triggering hypothesis was explored extensively by \citet{Clyne+1967,clyne1972perspectives,Clyne1980TriggeringAL,CLYNE+1987+739+764}, 
but these early investigations did not include any statistical analysis.
This was first introduced by \citet{broersma2006triggered}, who worked with ``a series of transcribed conversations between three Dutch-Moroccan Arabic bilinguals''.
This dataset was extremely small by modern standards (it included a few dozen switch points and a few dozen potential triggers). Similarly, \citet{broersma2009triggered} based her entire analysis on a single 24-minute interview with a single (Dutch--English speaking) informant.
Still, both were able to find statistically significant associations between triggering and CS.
More recently, \citet{soto2018role} extended this investigation to a larger corpus (the Bangor-Miami corpus of Spanish--English \citep{bangor-miami}), but focused only on a pre-defined list of cognates that they collected. 
In contrast, we work with much larger datasets that include thousands of switch points and potential triggers, in three different language pairs, and with both spoken dialogues and written social-media interactions.

\citet{broersma2006triggered} (and, subsequently, also \citet{broersma2009triggered} and \citet{soto2018role}) used the $\chi^2$ test to measure the correspondence between triggering and CS. We use the same measure
(more precisely, \emph{Fisher's exact test}, whose significance does not rely on an approximation that is only exact in the limit); but we extend the analysis by considering not only the statistical significance of the test, as determined by its $p$-value, but also the magnitude of the association between categories as an indication of the strength of the phenomena we observe, as determined by \emph{relative risk} (also known as \emph{risk ratio}). This facilitates a much more nuanced analysis of the results.

\section{Goals}
\label{sec:goals}

Our main goal in this work is to explore the triggering hypothesis more closely, focusing on a class of lexical items that we expect to be shared across the multiple mental lexicons of the multilingual speaker. Extending previous research, we aim at addressing the association between such shared items and CS in multiple datasets reflecting three different language pairs (EN--AR, EN--DE and EN--ES)%
\footnote{We use \emph{AR} for Arabic, \emph{DE} for German, \emph{EN} for English, and \emph{ES} for Spanish.}
and two different modalities (spoken and written).

\subsection{\emph{Shared} lexical items}
\label{sec:shared}
\citet{safaa:shuly:2022}
defined \emph{shared} lexical items as named entities in one language that are not translated to the other, and consequently have a similar form in both languages. They also included terms that lack (or have rare) translation equivalents in the other language.

%%%%
Following \citet{denglisch},
we refine the definition of \emph{shared} items by reflecting also the language in which such terms originate. 
%We also distinguish them from items that look and sound similar across two languages.
Our motivation is the assumption that a word like \textnl{taco}, which originates in Spanish but is fully adopted by English, may trigger code-switching from English to Spanish but perhaps less so in the reverse direction. 

In addition, words in $L_1$ that do not have a commonly used translation equivalents in $L_2$, and are hence used in both languages (e.g., \textnl{taxi}, which is commonly used in many Arab-speaking communities) are not considered a code-switch themselves but may trigger code-switching.%
% Additionally, we expect a word like \textnl{Lebanon}, which is an adaptation to English of the Arabic \textnl{Lubnan}, to trigger CS from English to Arabic to a lesser extent than, say, \textnl{Bahrain}, which originates in Arabic but sounds identical in English.
\footnote{Another deviation from the scheme of \citet{safaa:shuly:2022} is that we treat named entities that are specific to a foreign language as words in that language. For example, \textnl{Lebanon}, which is an English-specific variant of the Arabic \textnl{lubnan}, is viewed as an English token.}
Specifically, we divide the \emph{shared} category to three subcategories, depending on the origin of the word.
\begin{compactdesc}
    \item [Shared English]
    Named entities shared between two lexicons that originate in English, including person names (e.g., \textnl{Johnson}), commercial entities (e.g., \textnl{Twitter}, \textnl{Seven Eleven}), and geographic names that contain English words (e.g., \textnl{Times Square}). Also included are English-originating cultural terms that are adopted by the other language (e.g., \textnl{taxi}, \textnl{film})
    and English acronyms used cross-culturally on social media (e.g., \textnl{lol}).
    \item [Shared Arabic/German/Spanish]
    Named entities shared between the two lexicons%
    \footnote{These categories are defined separately for each language pair. E.g., \emph{shared-Arabic} is defined only for the EN--AR datasets. The same holds for \emph{shared-Other}.} 
    whose origin is Arabic (e.g., \textnl{Salah}, \textnl{Bahrain}); German (e.g., \textnl{Merkel}, \textnl{Berlin}); or Spanish (e.g., \textnl{Carlita}, \textnl{Guatemala}).
    Also, culturally dependent terms originating in these three languages that do not have translations in English, e.g., Arabic \textnl{Ramadan}, German \textnl{schnitzel} or Spanish \textnl{taco}.
    This category includes also interjections that are identified with one of these languages, e.g., Spanish \textnl{jajajaja}; and acronyms that expand to those languages 
    (e.g., Spanish \textnl{PR} for \textnl{Puerto Rico} or German \textnl{NRW} for \textnl{Nordrhein-Westfalen}).
    \item [Shared Other] %all other \emph{shared} lexical items 
    Words and terms that are used in both languages, but are not clearly identified with either of them, including named entities or terms that originate in a third language (e.g., \textnl{Erdogan}, \textnl{Pikachu} or \textnl{pizza}); terms whose origin is English but that do not include strong English linguistic features (e.g., \textnl{iPod}); interjections that are commonly used in both languages (e.g., \textnl{oh} or \textnl{wow}); person names that are common in both languages (e.g., \textnl{Lily}, \textnl{Adam}); and geographical terms that originate in a third language and are written and pronounced similarly in both languages (e.g., \textnl{Vietnam}).

\end{compactdesc}   

It is important to note that the tagging is context dependent: much like named entities, shared items may have different readings (i.e., tags) depending on the context in which they occur. Consider the two examples below. The token \textnl{warda} is tagged as Arabic in~\exref{ex:literal}, but as shared-Arabic in~\exref{ex:ne}. Consequently, using lists of shared items (as was done in previous work, e.g., by~\citet{soto2018role}), is not a sufficient solution.

\ex\label{ex:literal}
\begingl
\gla Maynf3sh warda wahda tayep ! //
\glb {it doesn't work} flower one only !//
\glft `It doesn't work , only one flower!' //
\endgl
\xe

\ex~\label{ex:ne}
\begingl
\gla kan beydafe3 3an amr warda //
\glb was defend about Amr Warda//
\glft `He was defending Amr Warda' //
\endgl
\xe

% For simplicity, we refer to both shared and similar items as (potential) \emph{triggers} below.
%%%%
Finally, note that some shared items are multi-word, e.g., \textnl{amr warda} (a person name) in~\exref{ex:ne}. When all tokens have the same origin $L$, we label the item shared-$L$; we do the same also when some tokens are shared-$L$ and others are shared-Other. But if one token is shared-$L_1$ and the other is shared-$L_2$, we label each token differently. For example, we tag \textnl{Nueva York} as shared-Spanish followed by shared-English.

\subsection{Hypotheses}
We pose the following hypotheses:
\begin{enumerate}
\item
\emph{Shared} lexical items are associated with CS, i.e., they tend to co-occur in the same utterances.
This is the main hypothesis investigated intensively by \citeauthor{Clyne:2003}'s many works and by subsequent research, but we  define shared items somewhat differently here, not relying on predefined (or manually annotated) lists of cognates and proper names.
\item
Such tendencies are more pronounced when the trigger is closer to the switch point.
Previous work investigated ``adjacent words'', whereas we investigate shared words located up to~6 tokens from the switch point.
\item
Triggers that precede the switch point are more strongly associated with CS than those that are adjacent to them.
\citet{broersma2006triggered} explain that
the trigger can succeed the CS point because language planning does not always work linearly, and the choice of language for words is not necessarily aligned with the linear order of these words in a sentence.
They therefore search for ``basic clauses'' that contain both switches and trigger words, in any order. We do not define basic clauses, resorting instead to a fixed-length window around shared items. But we do check, separately, the case of shared items that precede the CS point, and those that occur on either side of the CS point.
We do not separately investigate potential triggers that \emph{follow} the CS point because we expect the association in such cases to be weak. %
%\footnote{We did run several experiments that validated this expectation, in all our datasets.}
We focus instead on triggers \emph{near} the switch, on either side of it, and compare this situation with triggers that strictly precede the switch.
\item
Terms that originate in language $L_1$ are more likely to trigger a switch from $L_2$ to $L_1$ than the other way round. 
Our rationale here stems from the assumption that shared-$L_1$ words may be more deeply rooted in the lexicon of $L_1$ than the lexicon of $L_2$, even if they are included in both; and hence are more likely to trigger switches \emph{to} $L_1$ than \emph{from} $L_1$.
\end{enumerate}
These hypotheses are based on a precise definition of what constitutes a CS point (detailed in \sref{sec:definition-of-CS}).
But first, we describe the datasets we use to investigate these hypotheses.

\section{Data}
\label{sec:data}
We use five different datasets, in three language pairs.
The texts are either transcribed dialogues (in the case of Bangor-Miami) or sequences of utterances that constitute a \emph{thread} (in the case of social media). We view a turn of a single author/speaker as a basic unit; if the dataset is not already tokenized, we segment turns to utterances and then to tokens using NLTK \citep{nltk-book}.
Each token is then associated with a language ID tag.

\paragraph{Arabic--English}
We used the English--Arabizi (Arabic written in the Roman alphabet) dataset compiled and released by \citet{safaa:shuly:2022}. This corpus includes social media posts from Reddit and Twitter; it contains 2,643 utterances that were manually annotated for language ID (at the word level), which were used to train a highly accurate classifier (the accuracy of identifying words in Arabizi and English was~95\%; identifying \emph{shared} items was only~84\% accurate, with a precision of~89\% and much lower recall).
The classifier was then used to automatically annotate additional utterances, resulting in a total of over~865,000 utterances that include CS between English and Arabizi.
Each word in this dataset is associated with a unique language ID: Arabizi, English, French,%
\footnote{We focused only on AR--EN here because the number of French words in the corpus, and consequently the number of French--Arabic CS, was limited.
See Table~\ref{tbl:stats-ar}.}
Arabic, Shared, or Other.

We re-annotated the manually annotated data according to our revised definition of shared items and then retrained the classifier and applied it to the entire dataset.
%
%\safa{We also union back the sentences of the same tweet or comment back (belongs to the same sentence ID) in each corpus } 
We then combined the manually and automatically annotated subsets of each dataset; this resulted in two coherent datasets, one with Reddit posts and the other with Twitter comments. We report results for each dataset separately because they reflect different genres.
Table~\ref{tbl:stats-ar} lists statistics for the two EN--AR datasets.%
\footnote{The percentages do not always sum up to~1 because the datasets may include tokens with other tags (punctuation, emoji, hashtags, etc.)
Additionally, the numbers of shared tokens are actually counts of shared \emph{items}: if an item is multi-word, it is counted only once.}

\begin{table}[hbt]
\begin{center}
{\small
\begin{tabular}{l@{\hskip 7pt}rr@{\hskip 7pt}rr}
\multicolumn{1}{c}{\textbf{}} & \multicolumn{1}{c}{\textbf{Reddit}} & \multicolumn{1}{c}{\textbf{\%}}
& \multicolumn{1}{c}{\textbf{Twitter}}
& \multicolumn{1}{c}{\textbf{\%}}\\\hline
  Utterances   & 205,397 & & 659,958 & \\\hline
  Tokens (total)  & 3,855,900 & & 5,340,658 & \\
  Arabizi   & 585,830 & 15.2 & 2,978,070 & 55.8 \\
  English   & 2,678,442 & 69.5 & 1,639,966 & 30.7 \\
  French   & 7,363 & 0.2 & 6,872 & 0.1 \\
  Shared-EN & 7,779 & 0.2 & 17,849  & 0.3 \\ 
  Shared-AR & 31,992 & 0.8 & 19,190 & 0.4 \\
  Shared-Other & 20,333 & 0.5 & 11,679 & 0.2 \\\hline
  CS (total) & 274,200 & & 471,334 & \\
  EN$\rightarrow$AR & 133,642 & 48.7 & 233,616 & 49.6 \\
  AR$\rightarrow$EN & 140,558 & 51.3 & 237,718 & 50.4 \\
\end{tabular}
}
\caption{Statistics of the EN--AR datasets.}
\label{tbl:stats-ar}
\end{center}
\end{table}

\paragraph{Spanish--English}
We used two Spanish--English
datasets: \href{https://biling.talkbank.org/access/Bangor/Miami.html}{\emph{Bangor-Miami (BM)}} \citep{bangor-miami}, a corpus of transcribed Spanish--English bilingual speech; and \emph{SentiMix} \citep{aguilar-etal-2020-lince}, a dataset that was created for investigating sentiment analysis in a code-switched environment \citep{patwa2020sentimix}. 

Both corpora include token-level manually annotated language ID tags, but they use different schemes and include ambiguous language tags for named entities and cross-lingual terms. We manually changed the language tags of such tokens to English, Spanish, or a sub-class of \emph{Shared}, according to the scheme of Section~\ref{sec:shared}, so that they are consistent throughout the corpus.%
\footnote{No classifier was used on these datasets.}
Table~\ref{tbl:stats-es} lists statistics for the two EN--ES datasets.%
\footnote{MIX is a category of words that combine morphemes from the two languages; due to the relatively low number of such items, we ignore them in this work.}

% BM contains 42,854 utterances, 3,936 CS instances (1,673 of English to Spanish CS instances, 2,263 Spanish to English and 17 intra-word CS instances), and 14,656 Shared tokens (of which 5,603 are ``Shared-English'', 2,053 are ``Shared-Spanish'' and 7,018 are ``Shared-Other'').

\begin{table}[hbt]
\begin{center}
{\small
\begin{tabular}{lrrrr}
\multicolumn{1}{c}{\textbf{}} & \multicolumn{1}{c}{\textbf{BM}} & \multicolumn{1}{c}{\textbf{\%}} & \multicolumn{1}{c}{\textbf{SentiMix}} & \multicolumn{1}{c}{\textbf{\%}} \\\hline
  Utterances   & 42,854 & & 12,193 & \\\hline
  Tokens (total)  & 277,963 & & 186,585 & \\
  English   & 171,791 & 61.8 & 41,290 & 22.1 \\
  Spanish   & 91,419 & 32.9 & 91,419 & 49.0 \\
  Shared-EN & 5,659 & 2.0 & 1,125 & 0.6 \\
  Shared-ES & 2,042 & 0.7 & 1,752 & 0.9 \\
  Shared-Other & 7,035 & 2.5 & 1,585 & 0.8 \\ 
  MIX & 17 & 0.0 & 17 & 0.0 \\\hline
  CS (total) & 3,923 & & 19,226 & \\
  EN$\rightarrow$ES & 1,669 & 42.5 & 8,864 & 46.1 \\
  ES$\rightarrow$EN & 2,254 & 57.5 & 10,362 & 53.9\\
\end{tabular}
}
\caption{Statistics of the EN--ES datasets.}
\label{tbl:stats-es}
\end{center}
\end{table}

\paragraph{German--English}
We used the Denglisch corpus of mixed English--German Reddit posts compiled and released by \citet{denglisch}, with its original (``collapsed'') annotations, which are consistent with our scheme. As in the case of Arabizi, a small subset of this corpus (4,200 sentences) was annotated manually, and the remainder (over 228,000 sentences) was tagged by a classifier trained on the manually annotated subset. 
The overall word-level accuracy of the classifier was 96.5\%, with excellent (97--98\%) accuracy for English and German tokens, and 60-66\% for shared items (again, with much higher precision than recall).
Table~\ref{tbl:stats-de} lists statistics for this dataset.

\begin{table}[hbt]
\begin{center}
{\small
\begin{tabular}{lrrr}
\multicolumn{1}{c}{\textbf{}} & \multicolumn{1}{c}{\textbf{German}} & \multicolumn{1}{c}{\textbf{\%}} \\\hline
  Utterances   & 36,524 & \\\hline
  Tokens (total)  & 5,429,970 & \\
  English   & 1,826,171 & 33.6 \\
  German   & 2,281,859 & 42.0\\
  Shared-EN & 26,363 & 0.5 \\
  Shared-DE & 24,874 & 0.5 \\
  Shared-Other & 52,529 & 1.0 \\ 
  MIX & 4,187 & 0.1 \\\hline
  CS (total) & 270,375 & \\
  EN$\rightarrow$DE & 134,478 & 49.7\\
  DE$\rightarrow$EN & 135,897 & 50.3\\
\end{tabular}
}
\caption{Statistics of the EN--DE dataset.}
\label{tbl:stats-de}
\end{center}
\end{table}

Table~\ref{tbl:examples} depicts a few examples of utterances from our datasets, along with their annotation according to the scheme outlined above. 
\exref{ex:AR-EN} starts in English but then \textnl{Ahly} (an Egyptian football club) is mentioned; this token is tagged \emph{shared-Arabic}, and indeed after a few more English tokens, the author switches to Arabic.
In \exref{ex:ES-EN} the reverse pattern is observed: the utterance begins in Spanish, but two proper names tagged as \emph{shared-English} are introduced, and evidently the author switches to English. 
Finally, \exref{ex:DE-EN} begins in German and ends in English, perhaps in connection with the use of \textnl{schnitzel}, which is \emph{shared-German}.

\begin{table*}[hbt]
%\flushleft{\textbf{AR-EN}}
\ex~\label{ex:AR-EN}
\begingl
\gla every time I watch an ahly game I get goosebumps fel de2i2a el 74 //
\glb {} {} {} {} {} Ahly {} {} {} {} in minute the {}//
\glc EN EN EN EN EN SH-AR EN EN EN EN AR AR AR Other//
\glft `Every time I watch an Ahly game I get goosebumps in the 74th minute' //
\endgl
\xe

%\flushleft{\textbf{ES-EN}}
\ex~\label{ex:ES-EN}
\begingl
\gla tu eres scott y yo soy kourtney , had n't we agreed on this ? //
\glb you are Scott and I am Kourtney {} {} {} {} {} {} {} {} //
\glc ES  ES  SH-EN  ES  ES  ES  SH-EN  Other  EN  EN  EN  EN  EN  EN  Other//
\glft `You are Scott and I am Kourtney, hadn't we agreed on this?' //
\endgl
\xe

%\flushleft{\textbf{DE-EN}}
\ex~\label{ex:DE-EN}
\begingl
\gla Aba sie sagt ja making schnitzel for my husband //
\glb But she says yes {} {} {} {} {} //
\glc DE DE DE DE EN SH-DE EN EN EN//
\glft `But she says yes, making a schnitzel for my husband' //
\endgl
\xe

\caption{Example utterances with their language ID annotations.}
\label{tbl:examples}
\end{table*}

\section{Methodology}
\label{sec:methodology}

\subsection{Definition of CS points}
\label{sec:definition-of-CS}
To check the association between shared items and CS, the latter concept must be carefully defined, which is not always a trivial task \citep{alvarez-mellado-lignos-2022-borrowing}; previous work has sometimes been careless with this. 
We consider CS to be a property of a single token, defined as follows: A token $w$ is considered code-switched from $L_1$ to $L_2$ when:
\begin{inparaenum}[(i)]
    \item $w$ is labeled as $L_2$;
    \item it is preceded (in the same utterance) by a sequence of $n\ge 0$
    tokens labeled neither as $L_1$ nor as $L_2$; and
    \item this sequence is preceded (in the utterance) by a token labeled as $L_1$.
\end{inparaenum}
%
% Namely,
% \[\mathbf{utterance} = \cdot\cdot\cdot\tau\sigma_0\sigma_1\cdot\cdot\cdot\sigma_n\sigma\cdot\cdot\cdot\]
% \[l(\sigma)=L_2, l(\tau)=L_1, \forall i, l(\sigma_i) \notin {L_1,L_2}\]
This definition allows for sequences of shared lexical items (and other tokens, e.g., emoji) to intervene between a token in $L_1$ and a token in $L_2$; the CS point is the first $L_2$ token that follows such a sequence. 

Having said that, we exclude some CS points from our analysis: we treat \emph{insertional} switches differently from \emph{alternational} ones.  \citet{muysken:1997} defines alternation as ``a true switch from one language to the other, involving both grammar and lexicon''.  
All three examples in Table~\ref{tbl:examples} are alternational. In contrast, insertion is the embedding of a phrase from one language into an utterance that is otherwise in the other language.
%However, this definition treats \emph{insertional} and \emph{alternational} switches \citep{muysken:1997} in the same way. 
\exref{ex:insertional} demonstrates insertional CS: the English token \textnl{technically} is inserted into an otherwise fully Arabic utterance.

\ex~\label{ex:insertional}
\begingl
\gla Gama3a e7na technically fi ramadan //
\glb guys we're {} in Ramadan//
\glft `Guys, we're technically in Ramadan' //
\endgl
\xe

It is common to assume that insertional CS like the one in \exref{ex:insertional} involve \emph{a single} trigger, which affects the tendency to switch from $L_1$ to $L_2$; the switch back to $L_1$ is merely an inevitable consequence of the CS being insertional. Therefore, we exclude from our analyses the second switch in case of insertional CS.%
\footnote{We experimented also with the alternative approach, namely treating insertional and alternational switches identically. The results were pretty similar.}

We operationalize this as follows: given a sequence of tokens $w_1w_2w_3$, where $w_1$ and $w_3$ are in $L_1$ and $w_2$ is in $L_2$, we only consider $w_2$, but not $w_3$, to be a CS point. This does introduce noise occasionally, especially because some insertional switches involve the insertion of two, and sometimes even three tokens, as in \exref{ex:two-token-insertion}.
In such cases, the switch back to Arabic will (erroneously) be taken into consideration in our analyses.

\ex\label{ex:two-token-insertion}
\begingl
\gla Mafi good internet b kel lebnen //
\glb there-isn't {} {} in all Lebanon {}//
\glft `There's no good internet in all of Lebanon' //
\endgl
\xe

\subsection{Statistical analysis}
\label{sec:experiments}

To explore the associations between shared items (as defined in Section~\ref{sec:shared}) and CS (as defined above), we ran a multitude of statistical tests.  The tests vary in terms of the dataset used, the type of shared items investigated (the three sub-classes of shared items, or all shared items combined), the direction of the CS (from English to the other language or vice versa), whether the shared item \emph{precedes} the CS point or \emph{neighbors} it (given a shared item, we look for CS points following it, but also adjacent to it on either side), and the distance between the two (we look at CS distanced at most~1 to~6 tokens from the shared item). 

We now outline the structure of a single such test, where the dataset is the SentiMix corpus, the type of shared item is shared-English, the CS direction is EN$\rightarrow$ES, and the CS follows the item, at a distance of at most~2 tokens.   Table~\ref{tbl:contingency-table} depicts the data used in this test: it is a $2 \times 2$ contingency table, whose columns indicate if the lexical item is shared or not, and whose rows correspond to the presence or absence of CS points near the shared item. The sum of the numbers in the first column is the number of shared items in the dataset, and in the first row, the number of switch points investigated (a single CS point may be counted several times for different shared items). We exclude from the investigation the first and the last token in each utterance (the last token in an utterance cannot trigger a switch following it; and the first is limited in triggering a switch neighbouring it).

\begin{table}[hbt]
\centering
\begin{tabular}{llrr}
    &  & \multicolumn{2}{c}{\textbf{Is \emph{shared}}}\\ 
\textbf{\multirow{3}{*}{\rotatebox{90}{Near CS}}} &  & Yes & No\\\cline{2-4}
    & Yes  &  216   &  17515\\
    & No   &  659 &  143299\\\cline{2-4}
    &      &  24.7\% & 10.9\% \\
\multicolumn{4}{l}{Relative switching propensity: 2.266}\\
\multicolumn{4}{l}{$p$-value: $2.2\times 10^{-30}$}\\
\end{tabular}
\caption{Contingency table constructed for the SentiMix corpus, reflecting EN$\rightarrow$ES switches that follow shared-English lexical items at distance at most~2 from the switch point.}
\label{tbl:contingency-table}
\end{table}

Across the shared items in the dataset, the proportion of switches (within the specified distance) is $216/(216 + 659) = 24.7\%$, whereas across the non-shared items, this proportion is $17515/(17515 + 143299) = 10.9\%$.  Thus, the propensity to switch is $24.7 / 10.9 = 2.266$ times higher near a shared item, compared to a non-shared item.  We refer to the latter ratio as the \emph{relative switching propensity}; mathematically, it is analogous to the well known ``relative risk’’ (or ``risk ratio’’) from epidemiology and biostatistics \citep{rothman2012epidemiology}.  We test whether this ratio differs from~1 in a statistically significant way via a Fisher test, and obtain a $p$-value of $2.2\times 10^{-30}$, indicating a highly significant increased tendency to switch near a shared item.

Clearly, the relative switching propensity equals~1 if and only if the odds ratio equals~1, and the same statistical test (namely, Fisher’s) is appropriate for studying either of these quantities.  We prefer to use the relative switching propensity to quantify the magnitude of the association between shared items and CS, as it is more readily interpretable: in the above example, the number $2.266$ is our estimate for the factor by which switches are more common near shared items, compared to near non-shared items.

\section{Results}
\label{sec:results}

Section~\ref{sec:experiments} defines multiple statistical tests: first, we work with five datasets; for each dataset, we individually explore four types of shared items. For example, with Bangor-Miami we independently explore shared-English items, shared-Spanish, shared-Other, and all shared items combined.
We depict the results of each such ``multi-test'', with a specific dataset and a specific type of shared item, as one plot.

On each plot we depict the results of~36 statistical tests, which differ by the type of switch (EN$\rightarrow$ES or ES$\rightarrow$EN or both); whether the shared item precedes the CS point or neighbors it; and finally, the distance between the two (1--6). The result of each of these~36 statistical tests is depicted as a point in a graph, where the $X$ axis is the distance (1--6) and the $Y$ axis indicates the value of the relative switching propensity for the specific statistical test. This facilitates a clear view of the magnitude of the effect (i.e., the strength of the association) of each test. 
Additionally, when the $p$-value of a particular test is greater than~0.05
(the usual threshold for statistical significance), 
we indicate this as a black diamond marking on the point that corresponds to that test.

\begin{figure*}[hbt]
\centering
\includegraphics[width=\textwidth]{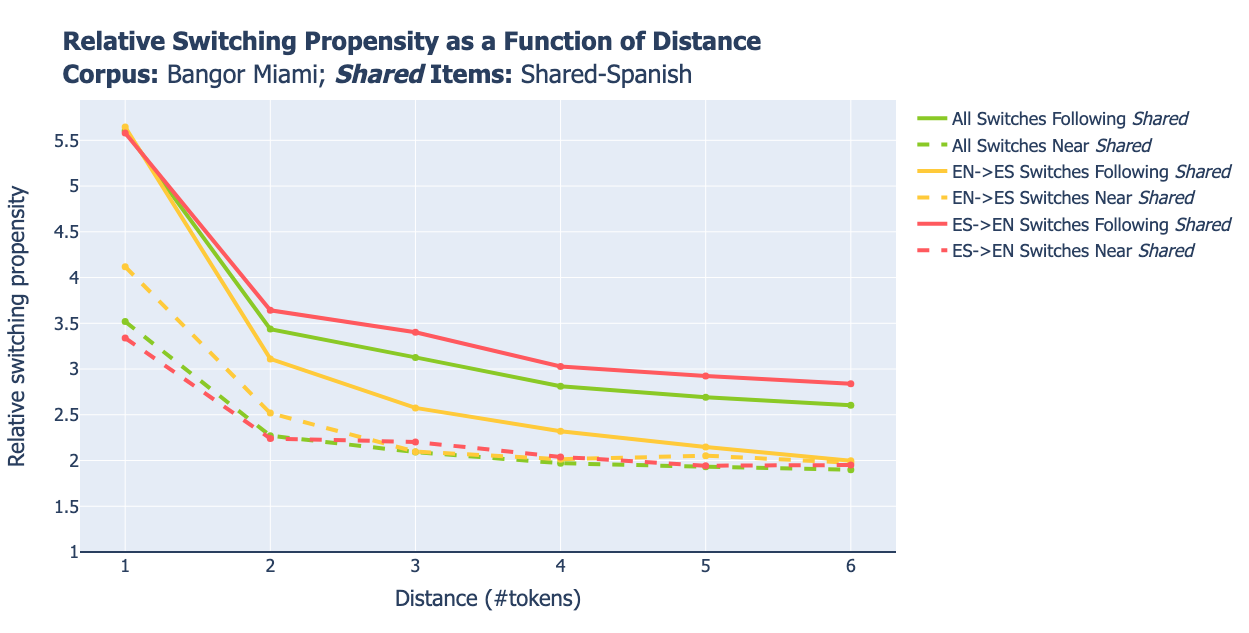}
\caption{A multi-test plot depicting the results of~36 tests on the Bangor-Miami corpus with shared Spanish items.}
\label{fig:example-plot}
\end{figure*}

Figure~\ref{fig:example-plot} shows the multi-test plot reflecting the results on the Bangor-Miami corpus with shared-Spanish items. The~36 points on this plot are connected by lines: solid lines reflect statistical tests where the shared item precedes the CS point, and dashed lines are for statistical tests where the shared item can occur before or after a CS point. The color of the line reflects the type of switch:
EN$\rightarrow$ES (yellow), ES$\rightarrow$EN (red), or both (green). 
See the legend to the right of the plot.

Several observations are revealed in Figure~\ref{fig:example-plot}. First and foremost, with no exceptions, all the tests yield statistically significant results ($p<0.05$), as there are no black diamonds on the plot. %
%\footnote{We indicate a data point that is not statistically significant by a black diamond, rather than a small dot in the color of the line. There is no such point in Figure~\ref{fig:example-plot}.}
This fundamentally supports our first hypothesis, namely that there is a clear association between shared items and CS. 
Furthermore, all the lines are monotonically decreasing, or at least non-increasing, thereby confirming our second hypothesis: the association between shared items and CS is stronger when the two are close, and diminishes as the distance between them increases. 

The fact that the solid lines are always above the dashed lines confirms our third hypothesis: the association is stronger with shared items that precede CS points than with shared items that are adjacent to them, on either side. 
Finally, the solid red line is always above the solid yellow line, indicating that shared-Spanish items are more strongly associated with CS from Spanish to English (red) than with CS from English to Spanish (yellow), in contrast to our hypothesis. 
This pattern, however, is partly reversed in the dashed lines: the jury is still out on our fourth hypothesis.

\begin{figure*}[hbt]
\centering
\includegraphics[width=\textwidth]{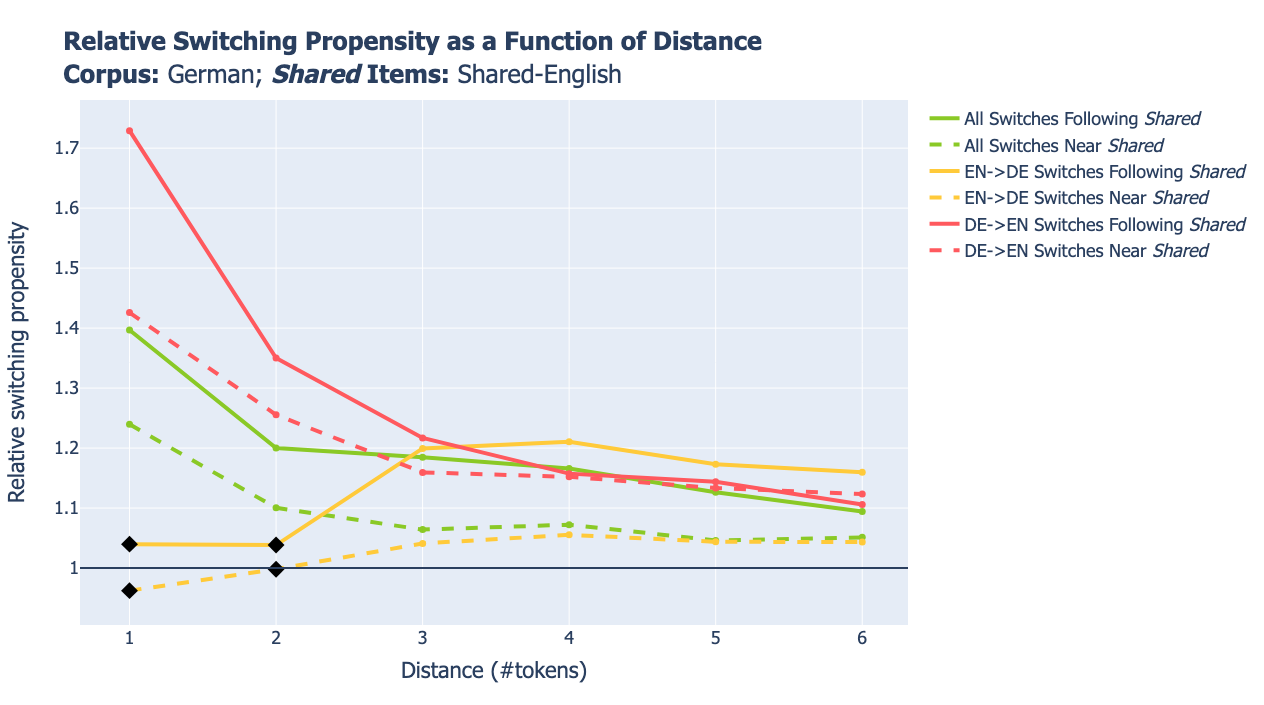}
\caption{A multi-test plot depicting the results of~36 tests on the EN-DE corpus with shared English items.}
\label{fig:denglisch}
\end{figure*}

While Figure~\ref{fig:example-plot} summarizes the results of~36 statistical tests, it is only one out of~20 similar ``multi-tests'': we have similar plots for five datasets, with four types of shared items per dataset. Space limitations prevent us from presenting all of them here (they will be included in the supplementary materials), but we do show a similar plot of the Denglisch EN-DE corpus, with data reflecting shared-English items, in Figure~\ref{fig:denglisch}. 
In addition, we now analyse the aggregate results of all~20 multi-tests in light of our four hypotheses.

\section{Analysis}
\label{sec:analysis}

\paragraph{Association between shared items and CS.}
Our first hypothesis was that shared items are indeed associated with CS. To assess the association, we expect the Fisher test to yield a statistically significant result in each of the statistical tests (i.e., no black diamonds in the plots). Not surprisingly, we do find such association.
Recall that each plot (such as the one in Figure~\ref{fig:example-plot}) depicts the results of~36 tests, and that we have 20 such plots. Of the~720 statistical tests, only~10 (1.4\%) yield $p$-values greater than~$0.05$.
We thus overwhelmingly establish the hypothesis that in all our datasets there is significant association between shared items of all kinds and CS, even when the shared item is as far as~6 tokens away from the CS point, and even when the shared item is adjacent to (i.e., may succeed) the CS point.
It is interesting to note that of the~10 exceptions,~8 are in the Denglisch corpus. We return to this below.

\paragraph{The impact of the distance between shared items and the CS point.}
Our second hypothesis was that the magnitude of the association diminishes as the shared item and the CS point are more distant.
To confirm this hypothesis we need to show that the lines in the plots are decreasing, or at least non-increasing. This is indeed the case for~98 (82\%) out of the~120 lines (6 per plot). Furthermore, most of the lines that are not decreasing include only a single point that violates the hypothesized trend. The main issue is, again, with the DE-EN dataset, which is responsible for~13 out of the~22 exceptions.

\paragraph{Shared items before or adjacent to CS points.}
We also hypothesized that the shared item ``triggers'' CS, namely that such items are more influential when they precede the CS point, as compared to when they are merely adjacent to it (on either side). To establish this hypothesis we need to show that the solid lines are above the dashed ones. 

Of the~720 data points in our~20 plots, only~38 did not comply with this condition; in other words, our hypothesis holds for almost~95\% of the points. 
One potential reason for the existence of outliers may be noise in our definition of insertional switches.
Recall from \sref{sec:definition-of-CS} that we try to find triggers for all switches except switches ``back from'' an insertion, but our definition of insertional CS assumes that they consist of exactly one token, whereas in reality some of them are multi-word expressions. 
We expect that switching ``out of'' such longer insertional switches does not require a trigger, but our analysis nonetheless looks for one.
Interestingly, all but one of the outliers involve switches \emph{from} English to the other language (yellow lines). We do not have an explanation for this observation.

\paragraph{The etymology of the shared item and its relation to the direction of the switch.}
Finally, we hypothesized that shared-$L_1$ items are more strongly associated with $L_2\rightarrow L_1$ switches than with $L_1\rightarrow L_2$ switches. This hypothesis is not supported by the data. For example, in the two AR-EN datasets, switches \emph{to} English were systematically more prominent than switches \emph{from} English, independently of the type of the shared item. 
In the EN-ES datasets, shared-EN and shared-ES items were associated with switches of both types almost to the same extent; and the DE-EN dataset also showed mixed, inconsistent results. 

One potential explanation for this observation has to do with insertional switches. With the exception of Bangor-Miami, our datasets reflect social media interactions on platforms that typically include discussions in English. We conjecture that when a particular discussion is conducted in another languages, insertions of English expressions are highly likely, much more than insertions of phrases in the other language to an otherwise English utterance.  
As mentioned above, our handling of insertional switches is noisy, which might affect the results.

A more theoretically based explanation of our failure to confirm the fourth hypothesis is grounded in theories of bilingualism which maintain that the two languages of a bilingual are both active simultaneously, and one of them has to be suppressed in order to yield words in the other \citep[e.g.,~][]{finkbeiner_gollan_caramazza_2006}. 
If this is indeed the case, then the origin of a shared item does not have to influence its likelihood to trigger a switch in any particular direction.

\paragraph{Summary}
Previous work on the triggering hypothesis focused solely on spoken dialogues and, consequently, was limited by the data available: typically, a few dozen dialogues spoken by a handful of participants in a single language pair. The extension to written CS, exemplified here, opens the door to investigations with vast amounts of data, but also raises interesting questions on the differences between written and spoken language and how CS is manifested in both modalities. Another interesting question has to do with the differences in the ways CS is manifested in closely-related language pairs (English--German, and to a lesser extent also English--Spanish) vs.\ in typologically unrelated language pairs (English--Arabic). A third dimension of comparison involves the differences in how CS is related to the status of the two languages involved: whether one of them is a minority language, a heritage language, or a lingua franca.  

A thorough investigation of all these issues is beyond the scope of this work; but we do note that among the five datasets used in this research, we did not find major differences between the (spoken dialogue) Bangor-Miami corpus and the (Twitter) SentiMix corpus, which are both in English--Spanish. This suggests that CS in written language, at least as it is used on very informal social media outlets, behaves similarly to CS in spoken language with respect to the triggering hypothesis.

We \emph{did} find significant differences between the Denglisch dataset and all others. Like the Denglisch corpus, one of the Arabizi datasets we studied also consists of Reddit posts, so the peculiarity of the English--German corpus cannot be attributed to the source of the texts it includes. 
As \citet{denglisch} note, this dataset is different from most corpora of bilingual language: German is the official language of Germany, where English is widely understood but is not a minority or heritage language, nor a lingua franca of a sub-community. This may result in a unique pattern of CS, different from the one observed in other language pairs, and might explain the  different results we obtain on this dataset.
We conjecture that CS between English and German is special because of the status of the two languages in German-speaking countries, but more research is needed to confirm this conjecture.

\section{Conclusion}
\label{sec:conclusion}
We investigated the triggering hypothesis using five datasets that reflect bilingual interactions in three language pairs. Employing standard yet powerful statistical methodology, we strongly confirmed three hypotheses: 
\begin{inparaenum}[(1)]
\item that there is a strong association between  code-switching and shared lexical items (proper names, but also culturally specific items that may lack translation equivalents in the other language);
\item that this association is stronger when the shared item precedes the switch point, rather than neighbors it; and
\item that the association diminishes as the shared item is farther away from the CS point.
\end{inparaenum}

We were unable to confirm a fourth hypothesis, namely that shared items originating in language $L_1$ are more likely to trigger a switch from $L_2$ to $L_1$ than the other way round. We do not know whether this is due to noise in our datasets or a bona fide property of bilingual language, rooted in cognitive-theoretical explanations; we leave this for future investigation. 

While the data used to establish the above results are unprecedented in terms of their size and diversity, at least in the psycholinguistic literature, we believe that they do not tell a full story. We would very much like to extend our datasets to more language pairs; to have sufficiently large datasets that would facilitate a comparative analysis of spoken vs.\ written data; and also enough data to compare CS between etymologically close languages vs.\ unrelated language pairs. We leave such investigations for future research.

\section*{Ethical considerations}
%This research was approved by the University of Haifa IRB. 
This research was approved by the University of Haifa IRB. 
We used previously collected data 
that are freely available for research purposes, and redistribute those data according to their original licenses.
All data are anonymized and we anticipate very minimal risk of abuse or dual use of the data.

\section*{Limitations}
Our datasets are by no means representative, and any conclusion resulting from their processing is limited to the population of speakers they reflect. 
However, the magnitude of the data we used here, especially compared to the sizes of corpora used previously to derive theories of code-switching, is sufficient to guarantee the replicability of our findings on further data.

\paragraph{Acknowledgements}
We are grateful to Melinda Fricke and Anat Prior for many discussions related to this work, and excellent ideas. We also thank the three  anonymous TACL reviewers for their valuable feedback and suggestions. This work was supported in part by grant No.\  2019785 from the United States-Israel Binational Science Foundation (BSF), and by grants No.\ 2007960, 2007656, 2125201 and 2040926 from the United States National Science Foundation (NSF).

%\clearpage
\bibliography{CodeSwitching,all}
\bibliographystyle{acl_natbib}

\end{document}